\documentclass[10pt,twocolumn,letterpaper]{article}

\usepackage{wacv}
\usepackage{times}
\usepackage{epsfig}
\usepackage{graphicx}
\usepackage{amsmath}
\usepackage{amssymb}

\def\wacvPaperID{43} 

\wacvfinalcopy 

\ifwacvfinal
\def\assignedStartPage{9876} 
\fi

\ifwacvfinal
\usepackage[breaklinks=true,bookmarks=false]{hyperref}
\else
\usepackage[pagebackref=true,breaklinks=true,colorlinks,bookmarks=false]{hyperref}
\fi

\ifwacvfinal
\else
\fi

\usepackage{times}
\usepackage{epsfig}
\usepackage{graphicx}
\usepackage{amsmath}
\usepackage{bbm}

\usepackage{amssymb}
\usepackage{caption}
\usepackage[nocompress]{cite}
\usepackage{slashbox}
\usepackage[dvipsnames,table]{xcolor} 
\usepackage{tabularx,ragged2e}
\usepackage{spreadtab}
\usepackage{adjustbox}
\usepackage{booktabs}
\usepackage{xcolor}
\usepackage{multirow}
\usepackage{rotating}
\usepackage{color, colortbl}
\usepackage{subfigure}

\usepackage[inline]{enumitem}
\usepackage{color,soul}

\graphicspath{{./figures/}}

\DeclareGraphicsExtensions{.jpeg,.png,.eps,.pdf}

\definecolor{Gray}{gray}{0.9}

\definecolor{Gray}{gray}{0.85}
\definecolor{LightCyan}{rgb}{0.88,1,1}

\newcolumntype{a}{>{\columncolor{Gray}}c}
\newcolumntype{b}{>{\columncolor{white}}c}

\newcommand{\partitle}[1]{\bigbreak\noindent\textbf{#1}}

\makeatletter
\newcommand*{\rom}[1]{\expandafter\@slowromancap\romannumeral #1@}
\makeatother

\usepackage{mathtools}

\DeclarePairedDelimiter\abs{\lvert}{\rvert}%

\usepackage{pifont}
\newcommand{\Ree}{\mathbb{R}}

\makeatletter
\newcommand*\bigcdot{\mathpalette\bigcdot@{.5}}
\newcommand*\bigcdot@[2]{\mathbin{\vcenter{\hbox{\scalebox{#2}{$\m@th#1\bullet$}}}}}
\makeatother

\definecolor{applegreen}{rgb}{0.55,0.71,0.0}
\definecolor{babyblue}{rgb}{0.54,0.81,0.94}
\definecolor{azure}{rgb}{0.0,0.5,1.0}
\definecolor{budgreen}{rgb}{0.48,0.71,0.38}
\definecolor{amaranthpurple}{rgb}{0.67,0.15,0.31}

\begin{document}

\title{Structured Visual Search via Composition-aware Learning}

\author{Mert Kilickaya, Arnold W.M. Smeulders\\
QUvA Lab, University of Amsterdam\\
{\tt\small kilickayamert@gmail.com, a.w.m.smeulders@uva.nl}
}

\maketitle

\begin{abstract}

This paper studies visual search using structured queries. The structure is in the form of a $2$D composition that encodes the position and the category of the objects. The transformation of the position and the category of the objects leads to a continuous-valued relationship between visual compositions, which carries highly beneficial information, although not leveraged by previous techniques. To that end, in this work, our goal is to leverage these continuous relationships by using the notion of symmetry in equivariance. Our model output is trained to change symmetrically with respect to the input transformations, leading to a sensitive feature space. Doing so leads to a highly efficient search technique, as our approach learns from fewer data using a smaller feature space. Experiments on two large-scale benchmarks of MS-COCO~\cite{mscoco} and HICO-DET~\cite{hicodet} demonstrates that our approach leads to a considerable gain in the performance against competing techniques. 

\end{abstract}

\section{Introduction}

\begin{figure}[t]
\begin{center}
  \includegraphics[width=\linewidth]{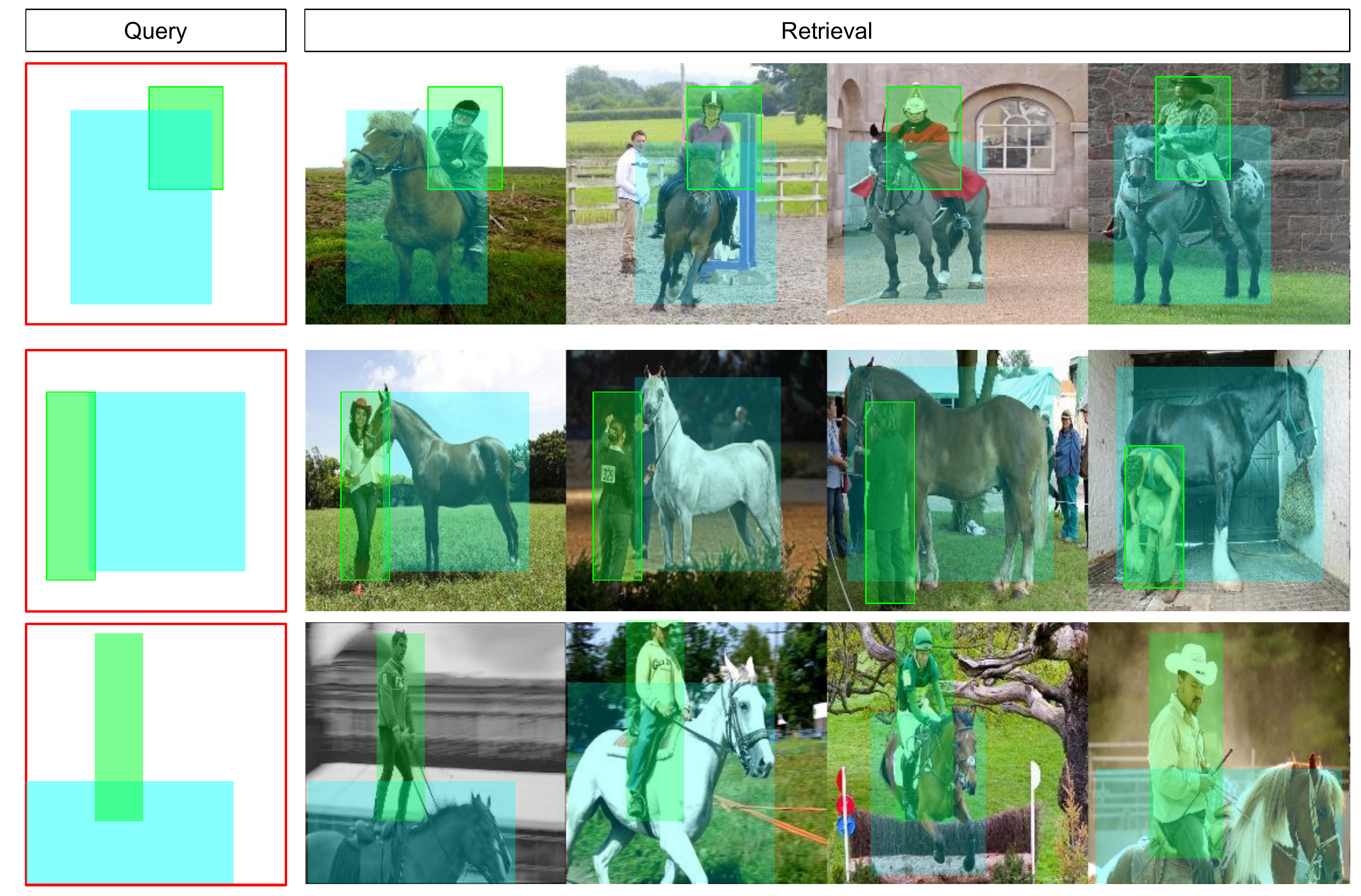}
  \caption{The compositional visual search takes a 2D canvas (left) as a query and then returns the relevant images that satisfy the object category and location constraints. Retrieval set (right) is in descending order by their mean Intersection-over-Union with the query canvas. Observe how small changes in the composition of the horse and the person lead to drastic transformations within the images. In this work, our goal is to learn these transformations for efficient compositional search.}
  \label{fig:teaser}
\end{center}
\end{figure}

Visual image search is a core problem in computer vision, with many applications, such as organizing photo albums~\cite{rodden2003people}, online shopping~\cite{jing2015visual}, or even in robotics~\cite{robot1,robot2}. Two popular means of searching for images are either text-to-image~\cite{im2text6,im2text2} or image-to-image~\cite{im2im9,im2im10}. While simple, text-based search could be limited in representing the \textit{intent} of the users, especially for the spatial interactions of objects. Image-based search can represent the spatial interactions, however, an exemplar query may not be available at hand. Due to these limitations, in our work, we focus on a structured visual search problem of compositional visual search.

The composition is one of the key elements in photography~\cite{composition}. It is the spatial arrangement of the objects within the image plane. Therefore, composition offers a natural way to interact with large image databases. For example, a big stock image company already offers tools for its users to find images from their databases by composing a query~\cite{shutterstock}. The users compose an abstract, 2D image query where they arrange the location and the category of the objects of interest, see Figure~\ref{fig:teaser}. 

Compositional visual search is initially tackled as a learning problem~\cite{comp2im}, recently using deep Convolutional Neural Networks (CNN)~\cite{vissynt}. Mai~\etal treats the problem as a visual feature synthesis task where they learn to map a given $2$D query canvas to a $3$ dimensional feature representation using binary metric learning which is then used for querying the database~\cite{vissynt}. We identify the following limitations with this approach: \textit{i)} The method requires a large-dimensional feature ($7\times7\times2048 \approx 100k$) to account for the positional and categorical information of the input objects, limiting the memory efficiency especially while searching across large databases. \textit{ii)} The method requires a large-scale dataset ($\approx 70k$ images) for training, limiting the sample efficiency. \textit{iii)} The method only considers binary relations between images, limiting the compositional-awareness. To overcome these limitations, in our work, we introduce composition-aware learning. 

\begin{figure}[t]
\begin{center}
  \includegraphics[width=1.\linewidth]{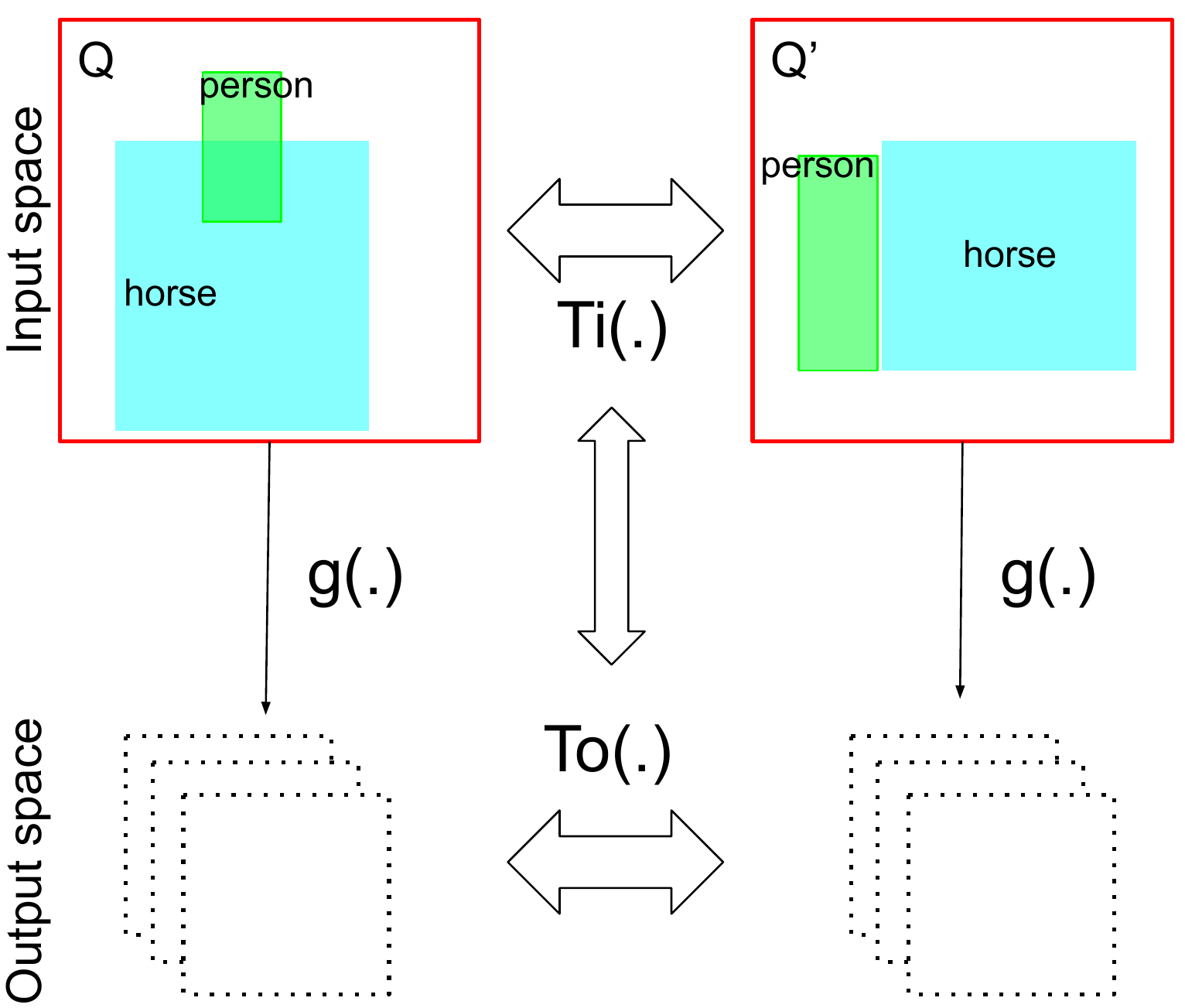}
  \caption{At the core of our technique is the principle of equivariance, which enforces a symmetrical change within the input and output spaces. We achieve this via mapping a query $Q$ and its transformed version $Q^{\prime} = T_{i}(Q)$ to a feature space where the transformation holds $g(Q^{\prime})= T_{o}(g(Q))$.}
  \label{fig:symmetry}
 \end{center}

\end{figure}

Compositional queries exhibit continuous-valued similarities between each other. Objects within the queries transform in two major ways: 1) Their positions change (translational transformation), 2) Their categories change (semantic transformation), see Figure~\ref{fig:teaser}. Our composition-aware learning approach takes advantage of such transformations using the principle of equivariance, see Figure~\ref{fig:symmetry}. Our formulation imposes the transformations within the input (query) space to have a symmetrical effect within the output (feature) space. To that end, we develop novel representations of the input and the output transformations, as well as a novel loss function to learn these transformations within a continuous range.

Our contributions are three-fold: 

\begin{enumerate}[label=\Roman*.]
    \item We introduce the concept of composition-aware learning for structured image search.
    \item We illustrate that our approach is efficient both in feature-space and data-space. 
    \item We benchmark our approach on two large-scale datasets of MS-COCO~\cite{mscoco} and HICO-DET~\cite{hicodet} against competitive techniques, showing considerable improvement. 
\end{enumerate}







\begin{figure*}[t]
\begin{center}
  \includegraphics[width=0.80\linewidth]{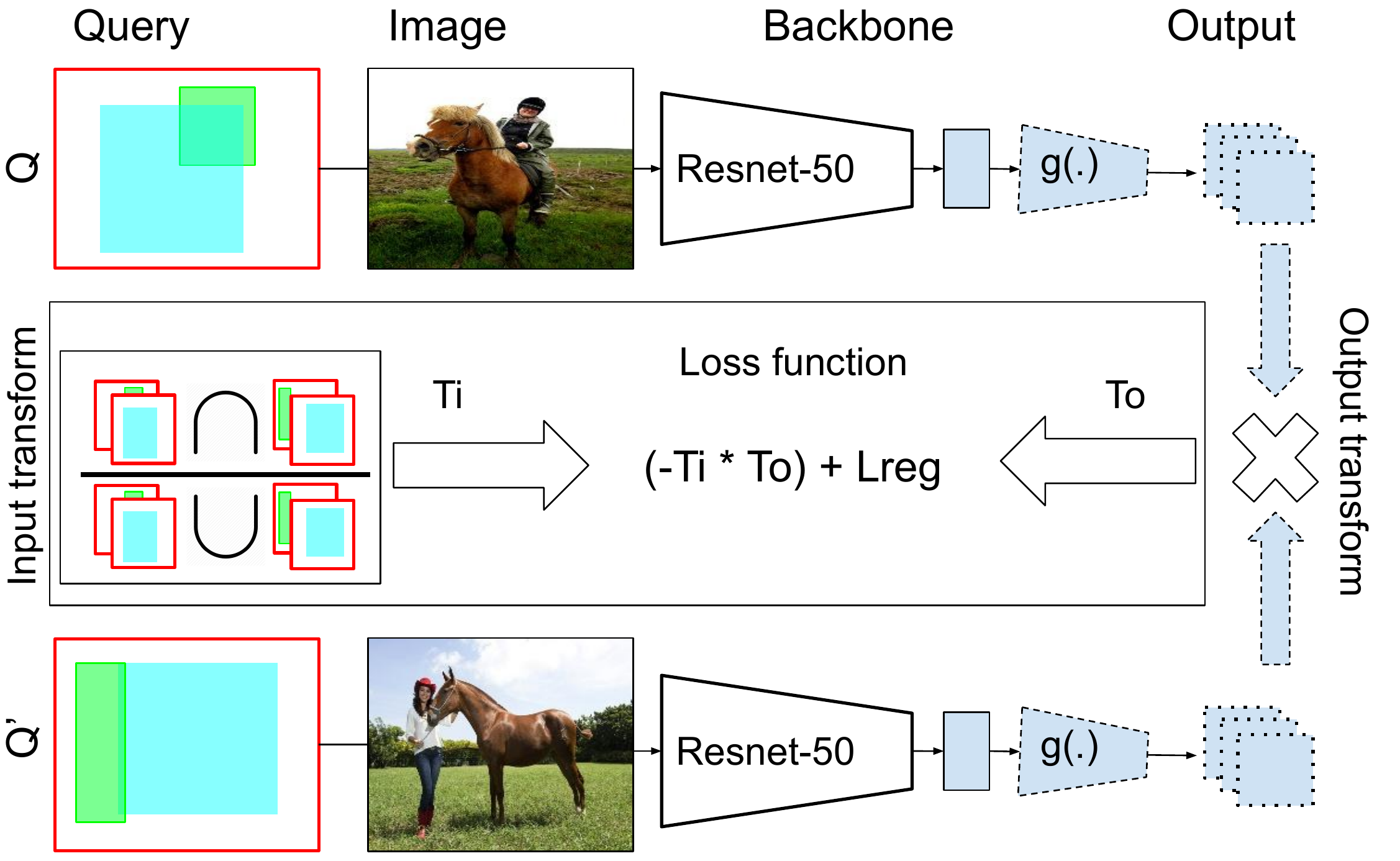}
\caption{Our composition-aware learning approach. Our approach is trained with pairs of queries $(Q, Q^{\prime})$ with identical backbones. 1) Given a pair of queries, we sample the corresponding images and feed them through a frozen ResNet-$50$ up to layer-$4$ of size $7\times7\times2048$. 2) Then, we process these activations with our light-weight $3$-layer CNN $g(\cdot)$ to map the channel dimension to a smaller size (\ie $2048\rightarrow256$) while preserving the spatial dimension of $7$. 3) In the mean-while, we compute the input ($T_{i}$) and the output ($T_{o}$) transformations, which are then forced to have similar values using the loss function.}
  \label{fig:approach}
 \end{center}
\end{figure*}

\section{Related Work}

\partitle{Compositional Visual Search.} Visual search mostly focused on text-to-image~\cite{im2text1,im2text2,im2text3,im2text4,im2text5,im2text6} or image-to-image~\cite{im2im1,im2im2,im2im3,im2im4,im2im5,im2im6,im2im7,im2im8,im2im9,im2im10,im2im11} search. Text-to-image is limited in representing the user intent, and a visual query may not be available for image-to-image search. Recent variants also combine the compositional query either with text~\cite{comp2im2} or image~\cite{comp2im4}. In this paper, we focus on compositional visual search~\cite{comp2im,comp2im3,vissynt}. A user composes an abstract, 2D query representing the objects, their categories, and relative locations which is then used to search over a potentially large database. A successful example is VisSynt~\cite{vissynt} where the authors treat the task as a visual feature synthesis problem using a triplet loss function. Such formulation is limited in the following ways: 1) VisSynt is high dimensional in feature-space ($100k$ dimensional), limiting memory efficiency, 2) VisSynt requires a large training set ($70k$ examples), limiting data efficiency, 3) VisSynt does not consider the compositional transformation between queries due to binary nature of the triplet loss~\cite{triplet}, limiting the generalization capability of the method. In our work, inspired by the equivariance principle, we propose composition-aware learning to overcome these limitations and test our efficiency and accuracy on two well-established benchmarks of MS-COCO~\cite{mscoco} and HICO-DET~\cite{hicodet}.



\partitle{Learning Equivariant Transformations.} Equivariance is the principle of the symmetry: 
Any change within the input space leads to a symmetrical change within the output space. Such formulation is highly beneficial, especially for model and data efficiency~\cite{efficiency}. In computer vision, equivariance is used to represent transformations such as object rotation~\cite{cohen2016group,equ1,equ2}, object translation~\cite{zhang2019making,equ6,equ7,equ8} or discrete motions~\cite{equivariance,jayaraman2016slow}. Our composition-aware learning approach is inspired by these works, as we align the continuous transformation between the input (query) and output (feature) spaces, see Figure~\ref{fig:symmetry}. 



\partitle{Continuous Metric Learning.} 
Continuous metric learning takes into account the continuous transformations between the image instances~\cite{kwak2016thin,contmetric2,logratio}, since such relationships can not be modeled with conventional metric learning techniques~\cite{contrastive,triplet}.
Recently, Kim \etal~\cite{logratio} proposed LogRatio, a loss function that matches the relative ratio of the input similarities with the output feature similarities. It yields significant gain over competing methods for pose and image caption search. Since compositional visual search is a continuous-valued problem, we bring LogRatio as a strong baseline to this problem. LogRatio intrinsically assumes a dense set of relevant images given an anchor point for an accurate estimation. However, compositional visual search follows Zipf distribution~\cite{zipf}, where, given a query, only a few images are relevant, limiting LogRatio performance.

\section{Composition-aware Learning}

\noindent
Our method consists of three building blocks: 

\begin{enumerate}[noitemsep,nolistsep]
\item Composition-aware transformation that computes the transformations in the input and output space, 
\item Composition-aware loss function that updates the network parameters according to the divergence of input-output transformations, 
\item Composition-equivariant CNN, used as the backbone to learn the transformation.
\end{enumerate}

\partitle{Method Overview.}
An overview of our method is provided in Figure~\ref{fig:approach}. Our method takes as an input a 2D compositional query $q \in \Ree^{H\times W}$, where $H, W$ are the height and width of the query canvas.
This query contains a set of objects, along with their categories and positions (in the form of bounding boxes).
The goal of our method is, given a target dataset of images, we want to retrieve the top-k images that are most relevant to the query $q$ -- \ie relevant to both the objects and their positions.
Each image $I$ can initially be represented as feature $x \in \Ree^{H^{\prime} \times W^{\prime} \times C^{\prime}}$ using the last convolutional layer of an off-the-shelf, ImageNet pre-trained deep CNN, \eg ResNet-$50$~\cite{resnet}.
Such feature $x$ preserves the spatial information as well as the object category information within the image $I$.
Furthermore, we assume access to a tuple $(c, x, I)$, where $c \in \Ree^{H \times W \times C}$ is a compositional map constructed using the object categories and bounding boxes of the query $q$. 
In addition, let $q^{\prime} = T(q)$ be the transformed version of the query $q$, and $(c^{\prime}, x^{\prime}, I^{\prime})$ are the corresponding composition map, CNN feature and the image. The transformation $T$ can correspond to a translation of object location(s), or a change in object categories in $q$. Our method trains a $3$-layer CNN $g_{\Theta}(\cdot)$ with the parameters $\Theta$, by minimizing the following objective function: 


\begin{align}
    \min_{\Theta} (L_{comp}(T_{i}(c, c^{\prime}), T_{o}(g_{\Theta}(x), g_{\Theta}(x^{\prime})))),
\end{align}

\noindent where $T_{i}$ measures the input transformation between compositional maps $c$ and $c^{\prime}$, and $T_{o}$ measures the transformation between output feature maps $g(x)$ and $g(x{\prime})$, and $L_{comp}$ is the composition-aware loss function measuring the discrepancy between these transformations. In the following, we first describe the compositional map $c$, and the input and the output transformations $T_{i}$ and $T_{o}$. Then, we describe composition-aware loss function $L_{comp}$. Finally, we describe our CNN architecture $g_{\Theta}(\cdot)$ that learns the mapping. We drop $\Theta$ from now for the sake of clarity.




\subsection{Composition-aware Transformation}


The goal of the composition-aware transformation is to quantify the amount of transformation between the input compositions $(c, c^{\prime})$ and output feature maps $(g(x), g(x^{\prime}))$ in the range $[0, 1]$. For this, first, we construct compositional maps from the input user queries, then we measure the input transformation using these maps, and finally we describe the output transformation. 

\partitle{Constructing compositional map $c$.} First, given a user query $q$ that reflects the category and the position of the objects, we create a one-hot binary feature map $c$ of size $\Ree^{H,W,C}$ where $[H,W]$ are the spatial dimension of the composition map ($H=W=32$), and $C$ is the number of object categories (\ie $80$ for MS-COCO~\cite{mscoco}). In this map, only the corresponding object locations and the categories are set to $1s$ and otherwise $0s$. This simple map encodes both the positional and categorical information of the input composition, which we will then use to measure the transformation within the input space. We apply the same procedure to the transformed query $q^{\prime}$ which yields $c^{\prime}$. Now given the pair of compositional maps $(c, c^{\prime})$, we can quantify the input transformation.


\partitle{Input transformation $T_{i}$.} Then, our goal is to measure the similarity between these two compositions as: 

\begin{align}
 \label{eq:input}
  T_{i}(c,c') =  \frac{ \sum_{xyz} (c_{xyz} \cdot c_{xyz}^{\prime}) }{\sum_{xyz}\mathbbm{1} (c_{xyz} + c_{xyz}^{\prime})},
\end{align}

\noindent where $\mathbbm{1}$ is an indicator function that is $1$ for only non-zero pixels. This simple expression captures the proportion of the intersection of the same-category object locations in the numerator and the union of the same-category object locations in the denominator. $T_{i}$ output is in the range $[0, 1]$, and will return $1$ if the two compositions $c$ and $c^{\prime}$ are identical in terms of object location and the categories, and $0$ if no objects share the same location. $T_{i}$ will smoothly change with the translation of the input objects in the compositions. Given the input transformation, we now need to compute the output transformation which will then be correlated with the changes within the input space. 



\partitle{Output transformation $T_{o}$.} Output transformation is computed as the dot product between the output features as follows: 

\begin{align}
    T_{o}(g(x), g(x^{\prime})) =  g(x) \times (g(x^{\prime}))^{\top},
\end{align}

\noindent where $(g(x^{\prime}))^{\top}$ is the transpose of the output feature $g(x^{\prime})$. We choose the dot product due to its simplicity and convenience in a visual search setting. $T_{o}$ can take arbitrary values in the range $[-\infty, \infty]$. In the following, we describe how to bound these values and measure the discepancy between the input-output transformations $T_{i}$ and $T_{o}$. 


\subsection{Composition-aware Loss}


Given the input-output transformations, we can now compute their discrepancy to update the parameters $\Theta$ of the network $g(\cdot)$. A \textit{naive} way to implement this would be to minimize the Euclidean distance between the input-output transformations as: 

\begin{align}
\label{eq:euclidean}
\min_{\Theta}\lVert \mathbf{T_{i} - \sigma(T_{o})} \rVert,
\end{align}

\noindent where $\sigma(\cdot)$ is the exponential non-linearity $\frac{1}{1 + \exp{(\cdot)}}$ to bound $T_{o}$ in range $[0, 1]$. However, such a function generates unbounded gradients therefore leading to instabilities during training~\cite{unbounded}, and reducing the performance, as we show through our experiments. Instead, cross entropy is a stable and widely used function that is used to update the network weights. However, cross entropy can only consider binary labels as $(0,1)$ whereas in our case the transformation values vary within $[0, 1]$. To that end, we derive a new loss function inspired by the cross entropy that can still consider in-between values. 

Consider that our goal is to maximize the correlation between input-output transformations as: 

\begin{align}
    \max_{\Theta} (T_{i} \cdot \sigma(T_{o}^{\top})).
\end{align}

We can also equivalently minimize the negative of this expression due to convenience: 

\begin{align}
    \min_{\Theta} ( - T_{i} \cdot \sigma(T_{o}^{\top})).
\end{align}


The divergence of $T_{o}$ and $T_{i}$ at the beginning of the training leads to instabilities during the training. To overcome this, we include additional regularization via the following two terms as:

\begin{align}
    \min_{\Theta} (T_{o} - T_{i} \cdot T_{o}^{\top} + \log(1 + \exp(-T_{o}))),
\end{align}

\noindent where the two terms $T_{o}$ and $\log(1 + \exp(-T_{o}))$ penalize for larger values of $T_{o}$
in the beginning of the training, leading to lesser divergence from $T_i$. To further avoid over-flow, the final form of the regularizer terms are: 

\begin{align}
    \min_{\Theta} (\max(T_{o}, 0.) - T_{i} \cdot T_{o}^{\top} + \log(1+\exp(-\lVert T_{o}\rVert))).
\end{align}

This is the final expression for $L_{comp}$ which we use throughout the training of our network $g(\cdot)$. 








\subsection{Composition-Equivariant Backbone}



Our model $g(\cdot)$ is a lightweight $3$-layers CNN that maps the bottleneck representation $x$ obtained from the pre-trained network ResNet-$50$ of dimension $\Ree^{7\times7\times2048}$ to a smaller channel dimension of the same spatial size, \ie $\Ree^{h\times w\times C}$, such as $7\times7\times256$ unless otherwise stated. Our intermediate convolutions are $2048\rightarrow1024\rightarrow512\rightarrow256$. The first two convolutions use $3\times3$ kernels whereas the last layer uses $1\times1$. We use stride$=1$ and apply zero-padding to preserve the spatial dimensions which are crucial for our task. We use $LeakyReLU$ with slope parameter $s=0.2$, batch-norm and dropout with $p=0.5$ in between layers. We do not apply any batch-norm, dropout, or $LeakyReLU$ at the output layer as this leads to inferior results. 

Since our goal is to preserve positional and categorical information, a network with standard layers may not be a proper fit. Convolution and pooling operations in standard networks are shown to be lacking translation (shift) equivariance, contrary to wide belief~\cite{zhang2019making}. To that end, we use the anti-aliasing trick suggested by~\cite{zhang2019making} to preserve shift equivariance throughout our network. Specifically, before computing each convolution, we apply a Gaussian blur on top of the feature map. This simple operation helps to keep translation information within the network layers.

\section{Experimental Setup}

\subsection{Datasets}

\partitle{Constructing Queries.} To evaluate our method objectively, without relying on user queries and studies, we rely on large-scale benchmarks with bounding box annotations. We evaluate our method on MS-COCO~\cite{mscoco} and HICO-Det~\cite{hicodet}. The training is only conducted on MS-COCO. Given an image, we select at most $6$ objects based on their area as is the best practice in~\cite{vissynt}. 


\partitle{MS-COCO.} MS-COCO is a large-scale object detection benchmark. It exhibits $80$ object categories such as animals (\ie dog, cat, zebra, horse) or house-hold objects. The dataset contains $120k$ training and $5k$ validation images. We split the training set to two mutually exclusive random sets of $50k$ training and $70k$ gallery images. The number of objects in each image differs in the range $[1, 6]$. 


\partitle{HICO-DET.} HICO-DET is a large-scale Human-object interaction detection benchmark~\cite{hicodet,kilickaya2020self}. HICO-DET builds upon $80$ MS-COCO object categories, and collects interactions for $117$ different verbs, such as ride, hold, eat or jump, for $600$ unique \texttt{<verb, noun>} combinations. Interactions exhibit fine-grained spatial configurations which makes it a challenging test for the compositional search. The dataset includes $37k$ training and $10k$ testing images. The training images are used as the gallery set and the testing set is used as the query set. A unique property of the dataset is that $150$ interactions have less than $10$ examples in the training set, which means a query can only match very few images within the gallery set, leading to a challenging visual search setup~\cite{kilickaya2020diagnosing}. HICO-DET is only used for evaluation. 

\subsection{Evaluation Metrics}


We evaluate the performance of the proposed model with three metrics. Standard mean Average Precision metric as is used in VisSynt~\cite{vissynt}. Also, we borrow continuous Normalized Discounted Cumulative Gain (cNDCG) and mean Relevance (mREL) metrics used in continuous metric learning literature~\cite{kwak2016thin,contmetric2,logratio} All metric values are based on the mean Intersection-over-Union (mIOU) scores between a query and all gallery images described below. For all three metrics, higher indicates better performance. 

\subsubsection{Mean Intersection-over-Union}

To measure the relevance between a query and a retrieved image, we resort to mean Intersection-over-Union as is the best practice~\cite{vissynt}. Concretely, to measure the relevance between a Query $q$ and a retrieved image $r$ 

\begin{equation}
    mIOU(q,r) = \frac{1}{\abs{B_{q}}} \sum_{b_{i} \in B_{Q}} \max_{b_{j} \in B_{I}} \mathbbm{1} (k(b_{i}) = k(b_{j})) \frac{b_{i} \cap b_{j}}{b_{i} \cup b_{j}},
\end{equation}

\noindent where $B_{Q}$ and $B_{I}$ represents all the available objects in the query $Q$ and retrieved image $I$ respectively, $\mathbbm{1}$ is an indicator function that checks whether objects $i$ and $j$ are from the same class $k$, which is then multiplied with the intersection-over-union between these two regions. This way, the metric measures both the spatial and semantic localization of the query object. 

\subsubsection{Metrics}

\partitle{mAP.} Based on the relevance score, we use mean Average Precision to measure the retrieval performance. We first use a heuristic relevance threshold $\geq0.30$ as recommended in~\cite{vissynt}, to convert continuous relevance values to discrete labels. Then, we measure the mAP values $@\{1, 10, 50\}$.

mAP metric does not respect the continuous nature of the compositional visual search since it binarizes continuous relevance values with a heuristic threshold. To that end, we resort to two additional metrics, continuous adaptation of NDCG and mean Relevance values which are used to evaluate continuous-valued metric learning techniques in~\cite{kwak2016thin,contmetric2,logratio}. 

\partitle{cNDCG.} We make use of the continuous adaptation of the Normalized Discounted Cumulative Gain as follows: 

\begin{equation}
    cNDCG(q) = \frac{1}{Z_{k}} \sum_{i=1}^{K} \frac{2^{r_{i}}}{\log_2(i+1)},
\end{equation}

\noindent that takes into account both the rank and the scores of the retrieved images and the ground truth relevance scores. In our experiments we report cNDCG$@\{1, 50, 100\}$. 

\partitle{mREL. } mREL measures the mean of the relevance scores of the retrieved images per query, which is then averaged over all queries. In our experiments, we report mREL$@\{1, 5, 20\}$. We also note the \textbf{oracle} performance where we assume access to the ground truth mIOU values to illustrate the upper bound in the performance. 

\subsection{Performance Comparison}

\partitle{ResNet-$50$~\cite{resnet}.} We use the activations from layer-$4$ of ResNet-$50$ to retrieve images. In this work, we build upon this feature since it captures the object semantics and positions within the feature map of size $\Ree^{7\times7\times2048}$. We also experimented with the earlier layers, however we found that layer-$4$ performs the best. The network is pre-trained on ImageNet~\cite{imagenet}. 

\partitle{Textual.} We assume access to the ground truth object labels for a query and retrieve images that contain the same set of objects. This acts as a textual query baseline and is blind to the spatial information. 

\partitle{VisSynt~\cite{vissynt}.} This baseline uses a triplet loss formulation coupled with a classification loss to perform a compositional visual search. We use the same backbone architecture $g(\cdot)$ and the same target feature ResNet-$50$ to train this baseline for a fair comparison.

\partitle{LogRatio~\cite{logratio}.} This method is the state-of-the-art technique in continuous metric learning, originally evaluated on human pose and image caption retrieval. In this work, we bring this technique as a strong baseline since the visual composition space also exhibits continuous relationships. We use the authors code~\footnote{https://github.com/tjddus9597/Beyond-Binary-Supervision-CVPR19} and the recommended setup. We convert mIOU scores to distance values as $1-mIOU$ since the method minimizes the distances. 

\partitle{Implementation details.} We use PyTorch~\cite{pytorch} to implement our method. We use the same backbone ($g(\cdot)$) and the input feature (ResNet-$50$) for all the baselines. All the models are trained for $20$ epochs using SGD with momentum ($=0.9$). We use an initial learning rate of $10^{-2}$ which is decayed exponentially with $0.004$ at every epoch. We use weight decay ($wd=0.005$) for regularization. In practice, we compute input-output transformations between all examples within the batch to get the best out of each batch. We set the batch size to $36$, and given each query in the batch, we sample $1$ highly relevant and $1$ less relevant examples for each query, which leads to an effective batch size of $36\times3=108$.

\begin{figure*}[t]
\minipage{0.29\textwidth}
  \includegraphics[width=\linewidth]{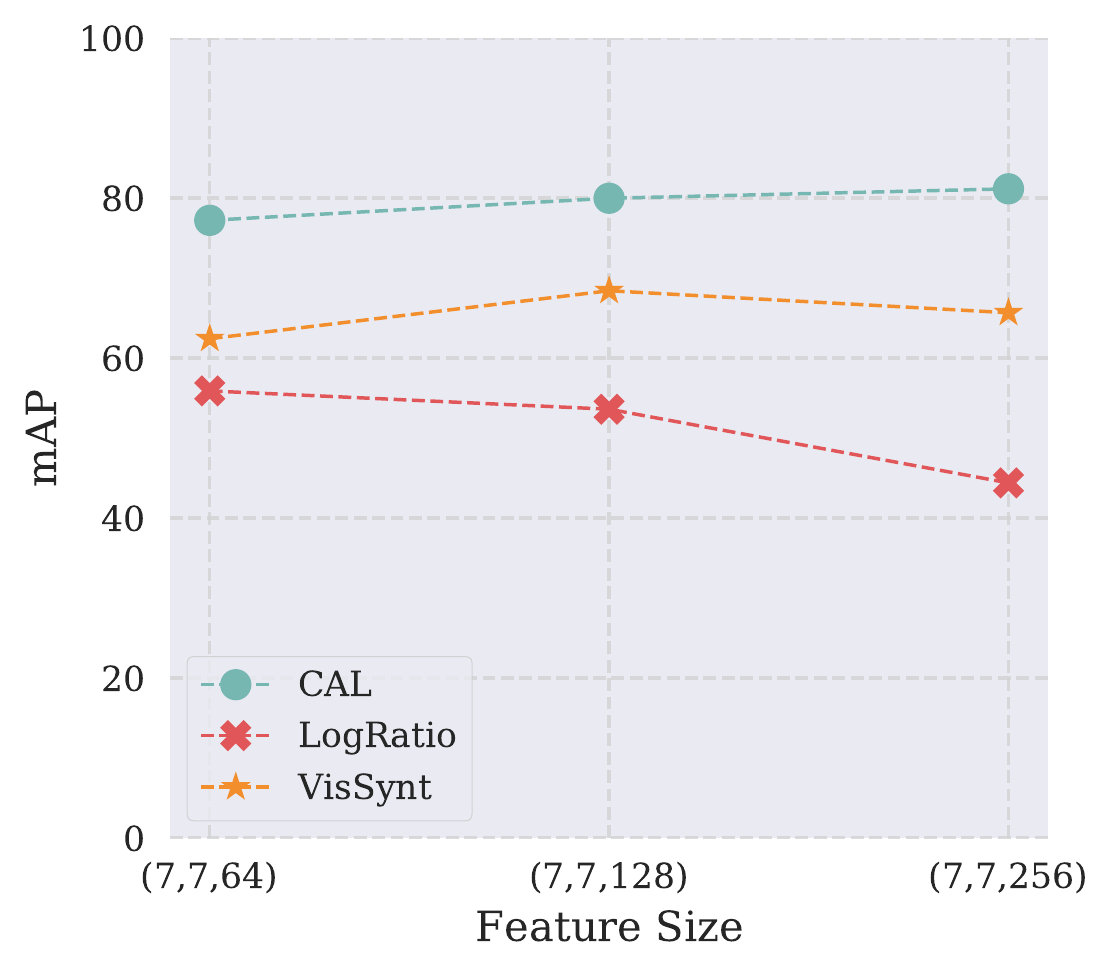}
\endminipage\hfill
\minipage{0.29\textwidth}
  \includegraphics[width=\linewidth]{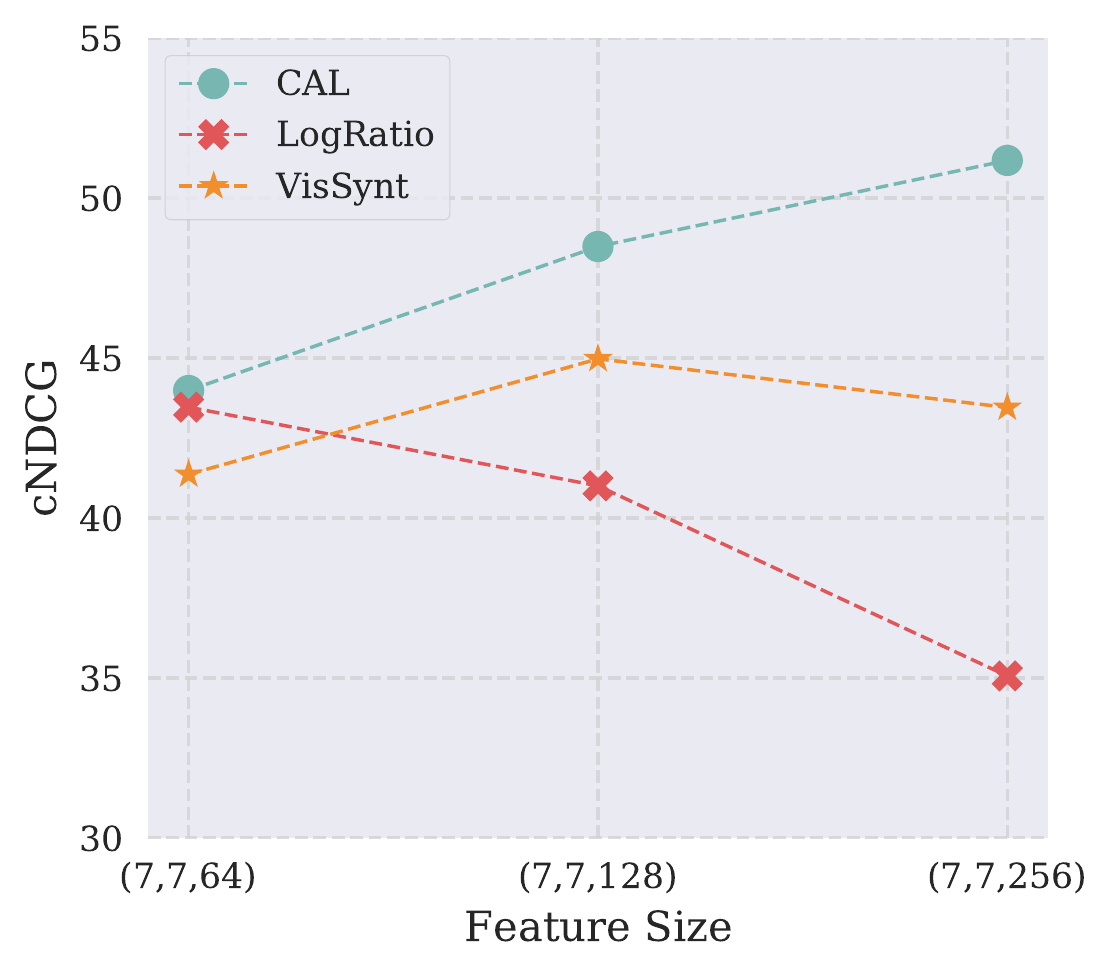}
\endminipage\hfill
\minipage{0.29\textwidth}%
  \includegraphics[width=\linewidth]{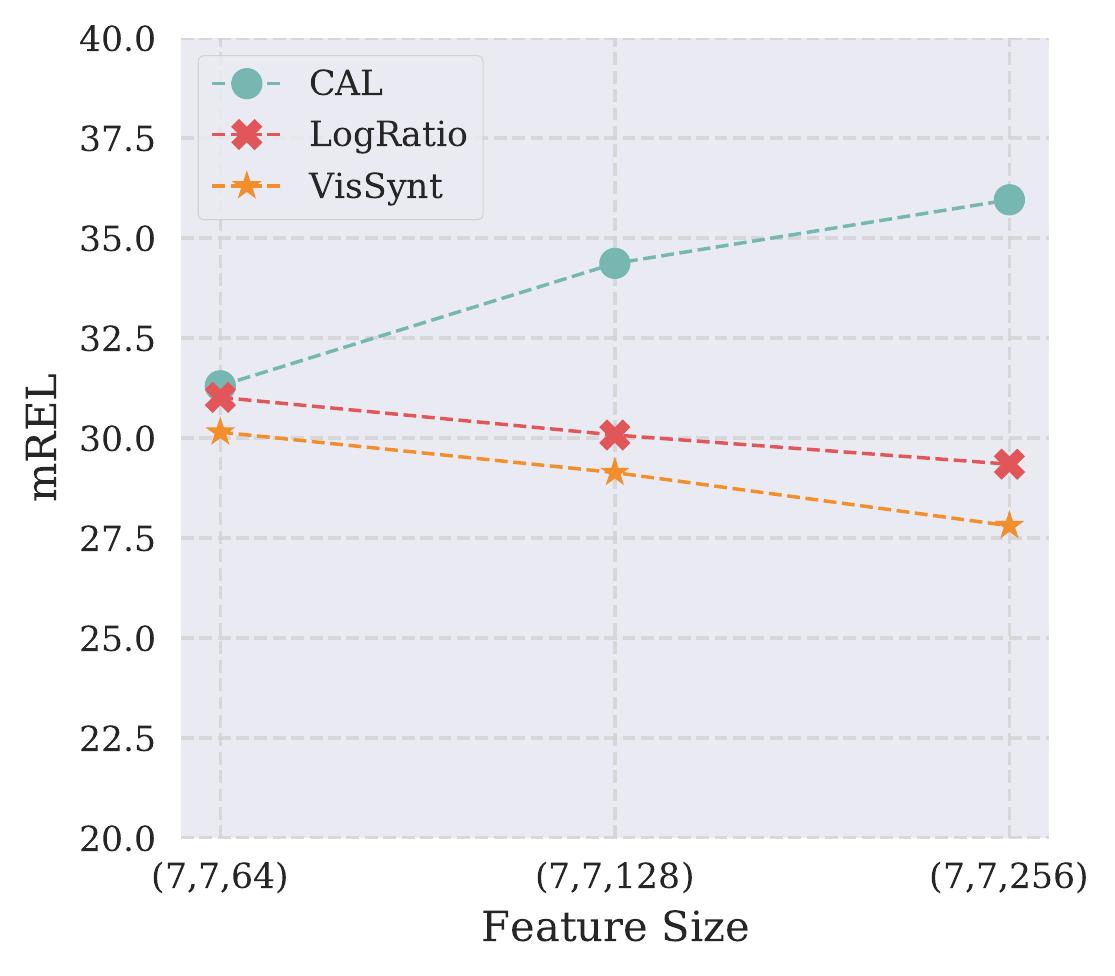}
\endminipage
\caption{Feature efficiency. Our model performs better even when the feature-space is compact.}
\label{fig:featureeff}
\end{figure*}

\begin{figure*}[t]
\minipage{0.29\textwidth}
  \includegraphics[width=\linewidth]{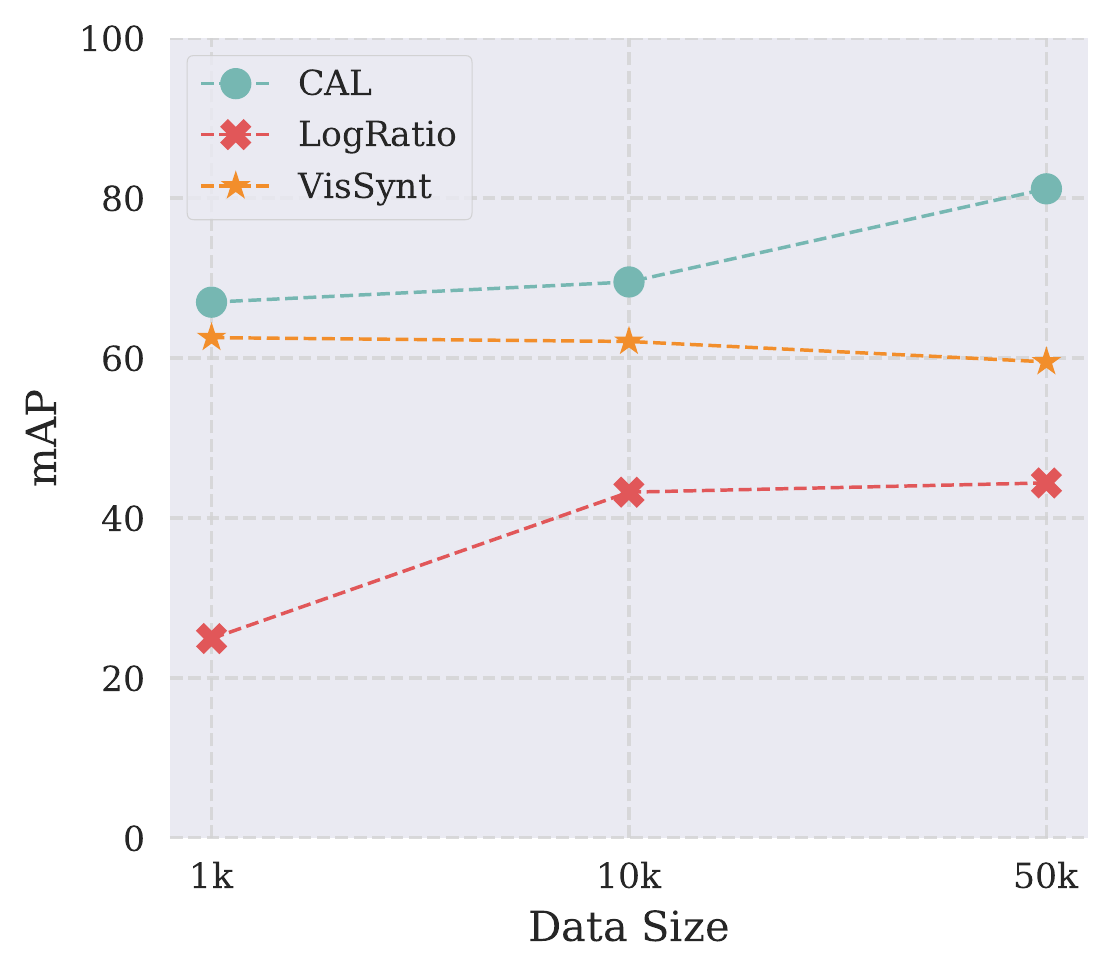}
\endminipage\hfill
\minipage{0.29\textwidth}
  \includegraphics[width=\linewidth]{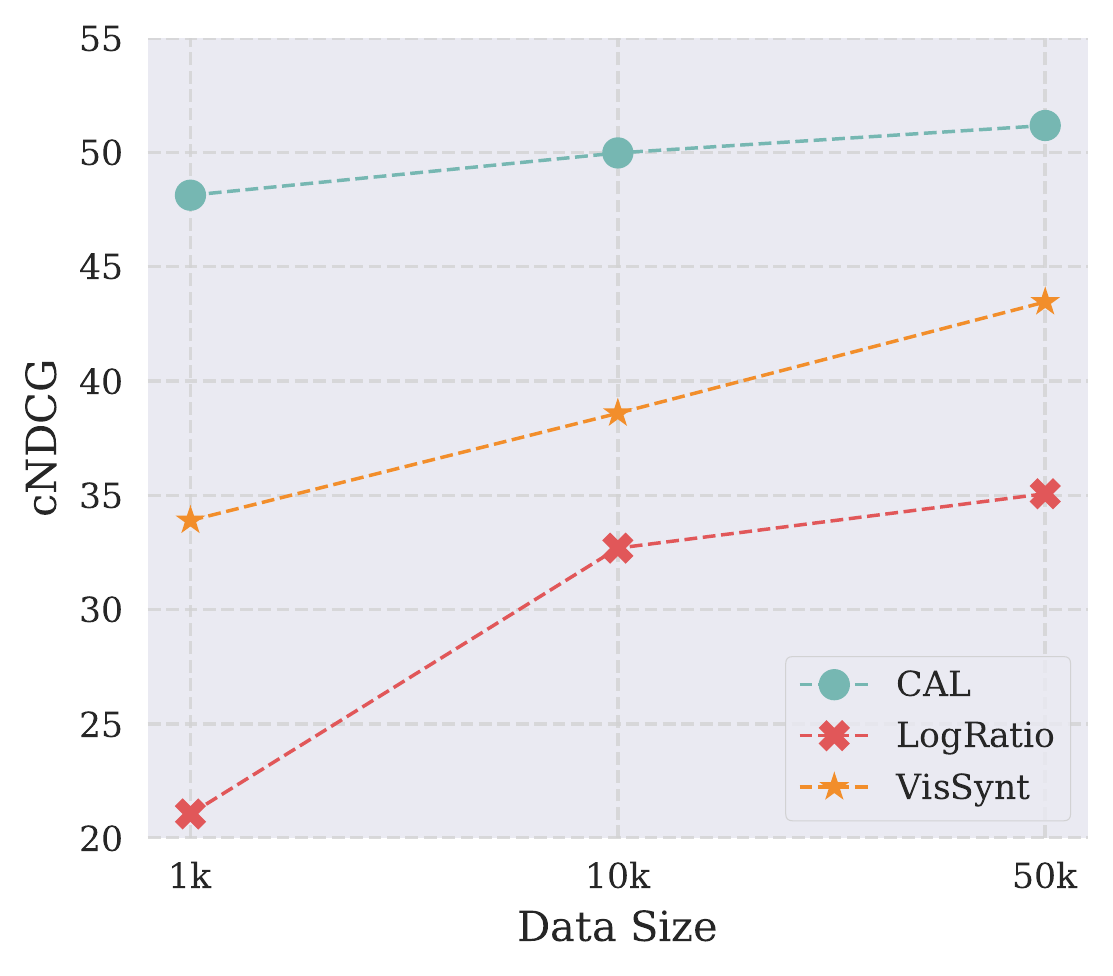}
\endminipage\hfill
\minipage{0.29\textwidth}%
  \includegraphics[width=\linewidth]{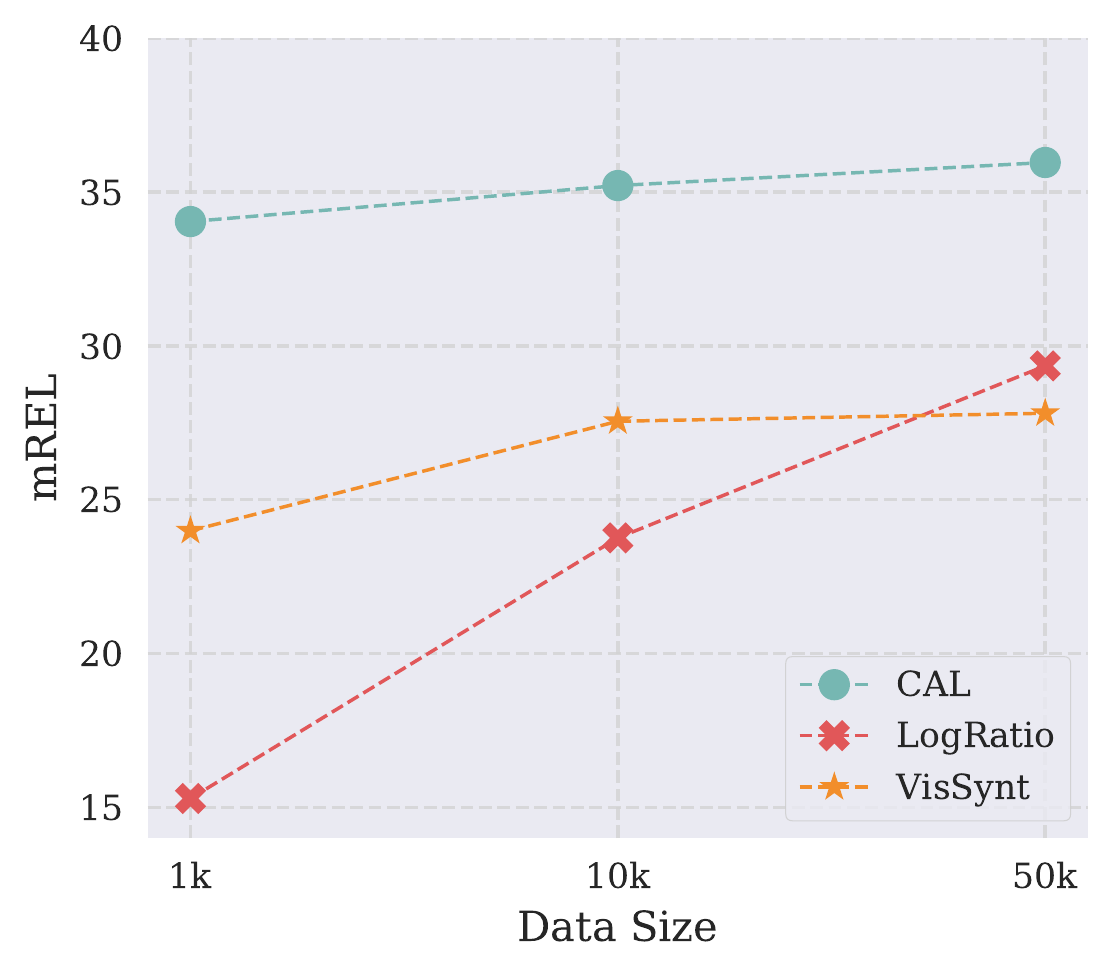}
\endminipage
\caption{Data efficiency. Our model outperforms VisSynt and LogRatio within small data regime.}
\label{fig:dataeff}
\end{figure*}

\section{Experiments}

In this Section, we present our experiments. For Experiments $1-2$, we use all three metrics $@k=1$. For the third experiment of the State-of-the-Art comparison, we provide performance at different $k$ values. 

\subsection{Ablation of Composition-aware Learning}


\partitle{Euclidean vs. Composition-aware loss.} In our first ablation study, we compare the Euclidean loss described in Equation~\ref{eq:euclidean} with our composition-aware loss. The results are presented in Table~\ref{tab:ablation}.

\begin{table}[h]
\begin{center}
\begin{tabular}{lccc}
\toprule 

& mAP & cNDCG & mREL \\ 
\midrule 

Euclidean & $66.87$ & $39.73$ & $28.49$ \\  
CAL (ours) & $81.17$ & $51.18$ & $35.96$ \\  

\bottomrule
\end{tabular}

\end{center}
\caption{Euclidean vs. Composition-aware loss.}
\label{tab:ablation}
\end{table}


It is observed that Composition-aware loss outperforms Euclidean alternative by a large-margin, confirming the effectiveness of the proposed loss function. 

\begin{table}[h]
\begin{center}
\begin{tabular}{lccc}
\toprule 

& mAP & cNDCG & mREL \\ 
\midrule 

Lingual & $65.14$ & $27.77$ & $19.56$ \\  
Visual (ours) & $81.17$ & $51.18$ & $35.96$ \\    

\bottomrule
\end{tabular}

\end{center}
\caption{Lingual vs. Visual input transformation.}
\label{tab:ablation2}
\end{table}

\partitle{Lingual vs. Visual transformation.} In our second ablation study, we test the domain of the input transformation (Eq~\ref{eq:input}). In our work, we proposed a visual-based input transformation whereas VisSynt~\cite{vissynt} utilizes a lingual-based input transformation using semantic Word2vec embeddings~\cite{word2vec}. As can be seen from Table~\ref{tab:ablation2}, vision-based transformation outperforms the lingual counterpart, since it can better encode the relationships within the visual world. 



\subsection{Feature and Data Efficiency}


In this experiment, we test the efficiency. Specifically, we first test the feature-space efficiency to see how the performance changes with varying sizes of the query embedding. Second, we test the data-space efficiency by sub-sampling the training data.


\partitle{Feature-space efficiency.} We change the feature embedding size by varying the number of channels as ${64,128,256}$ by keeping the spatial dimension of $7\times7$. We compare our approach to VisSynt~\cite{vissynt} and LogRatio~\cite{logratio}. The results can be seen from Figure~\ref{fig:featureeff}.

As can be seen, our approach performs the best for all metrics and across all feature sizes. This indicates that composition-aware learning is effective even when the feature size is compact (\ie $7\times7\times64$). Another observation is that the performance of $CAL$ increases with the increased feature size, whereas the performance of the two other techniques is lower. This indicates that $CAL$ can leverage bigger feature sizes while other objectives tend to over-fit. 

It is concluded that $CAL$ is a feature-efficient approach for compositional visual search.

\partitle{Data-space Efficiency.} In this experiment we vary the number of training data as ${1k, 10k, 50k}$. The results can be seen from Figure~\ref{fig:dataeff}.

Our method performs the best regardless of the training size. The gap in the performance is even more significant when the training set size is highly limited (\ie $1k$ only), confirming the data efficiency of the proposed approach. 

It is concluded that $CAL$ can learn more from fewer examples by leveraging the continuous-valued transformations and the regularized loss function.





\subsection{Comparison with the State-of-the-Art}

\begin{figure*}[t]
\minipage{0.28\textwidth}
  \includegraphics[width=\linewidth]{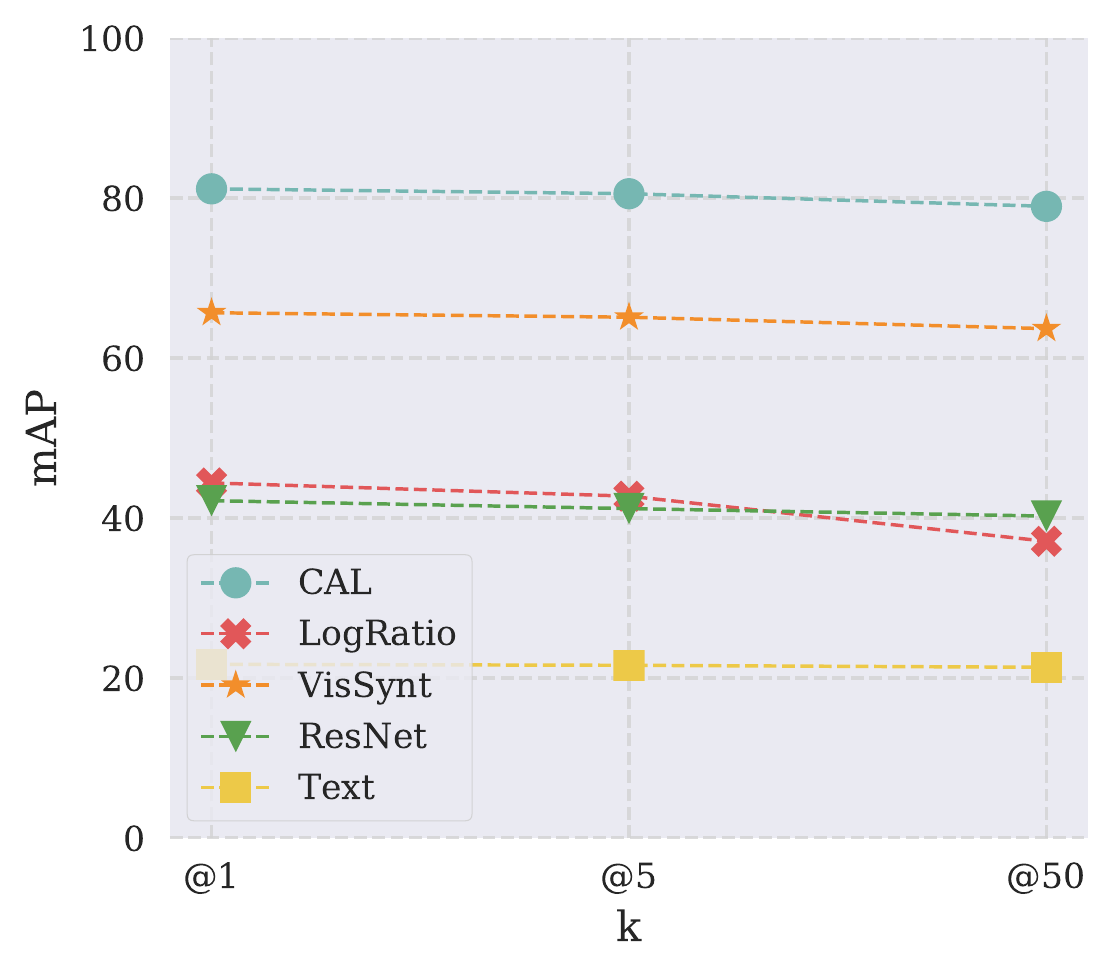}
\endminipage\hfill
\minipage{0.28\textwidth}
  \includegraphics[width=\linewidth]{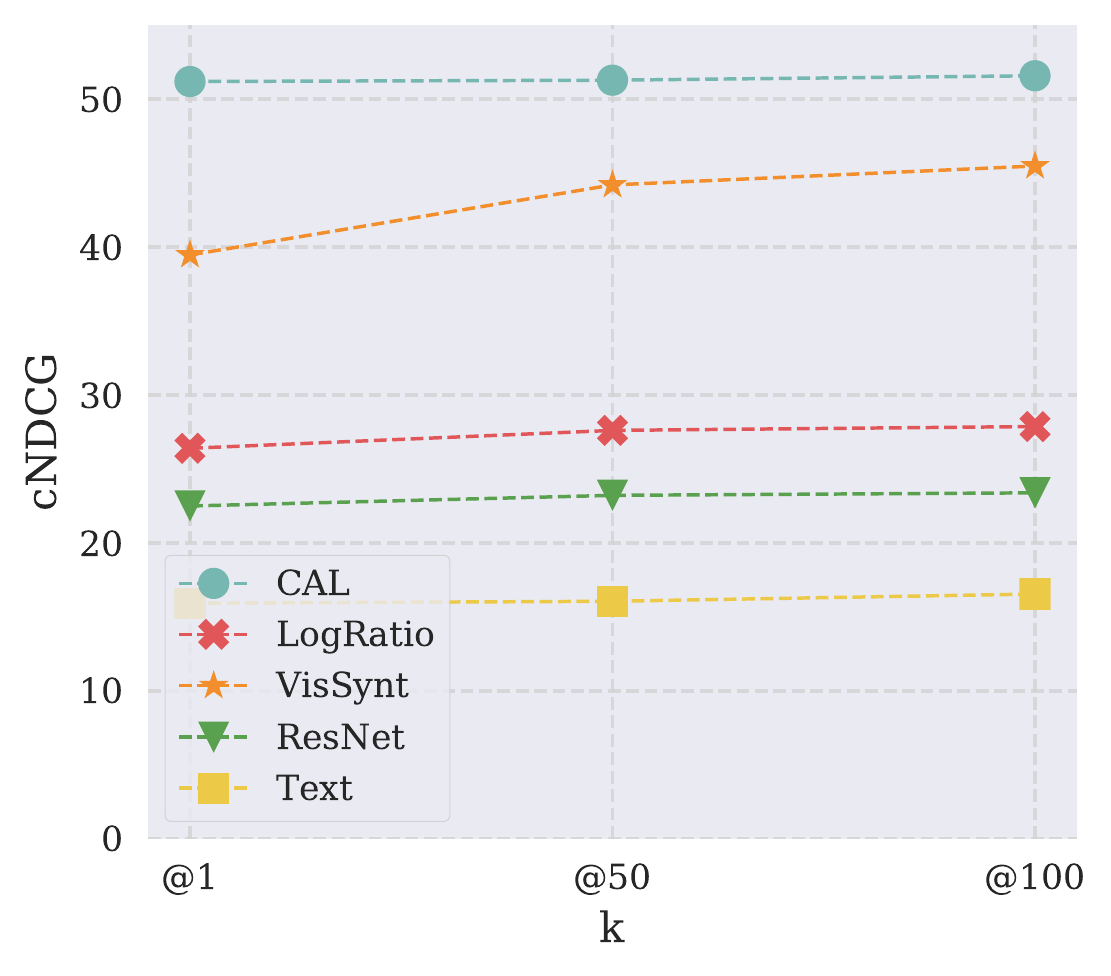}
\endminipage\hfill
\minipage{0.28\textwidth}%
  \includegraphics[width=\linewidth]{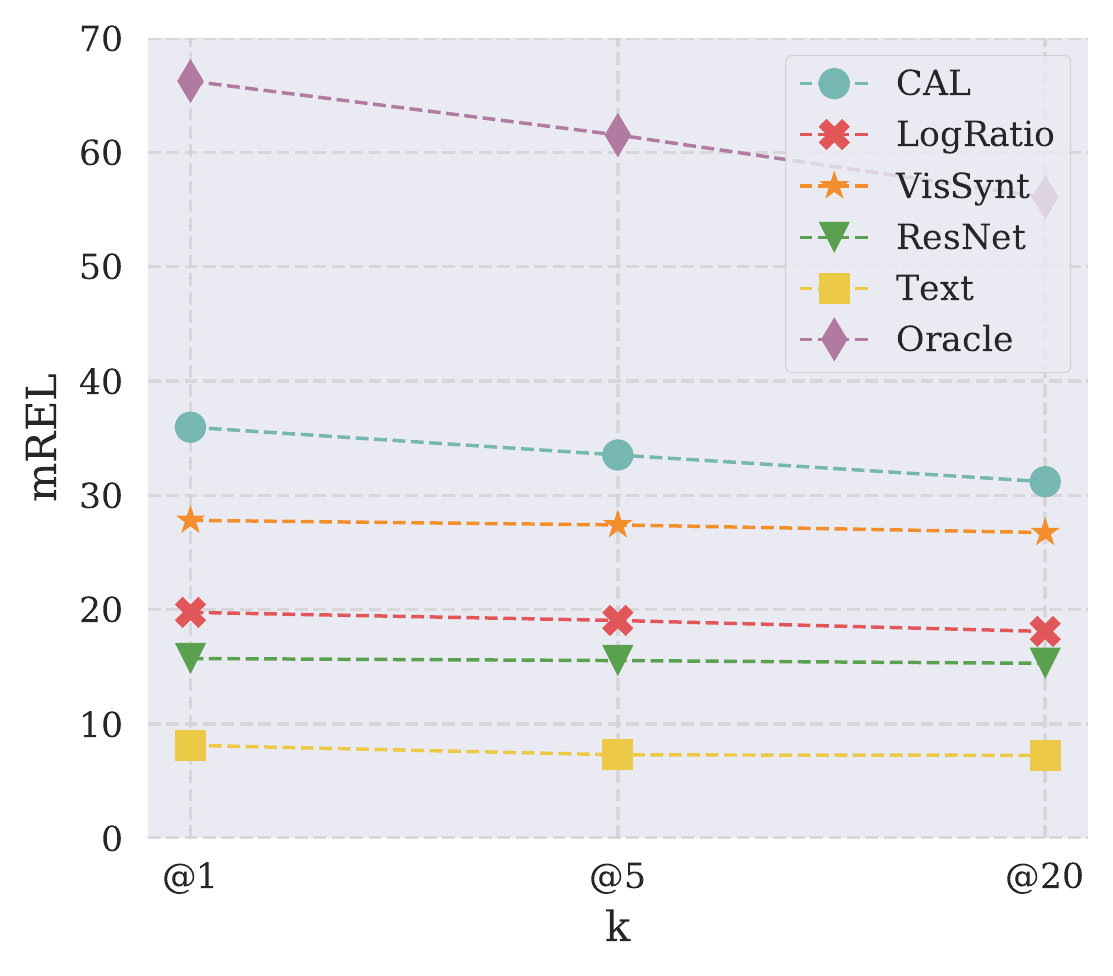}
\endminipage
\caption{Benchmarking on MS-COCO~\cite{mscoco}. Our method outperforms existing techniques for all three metrics.}
\label{fig:sotacoco}
\end{figure*}

\begin{figure*}[t]
\minipage{0.28\textwidth}
  \includegraphics[width=\linewidth]{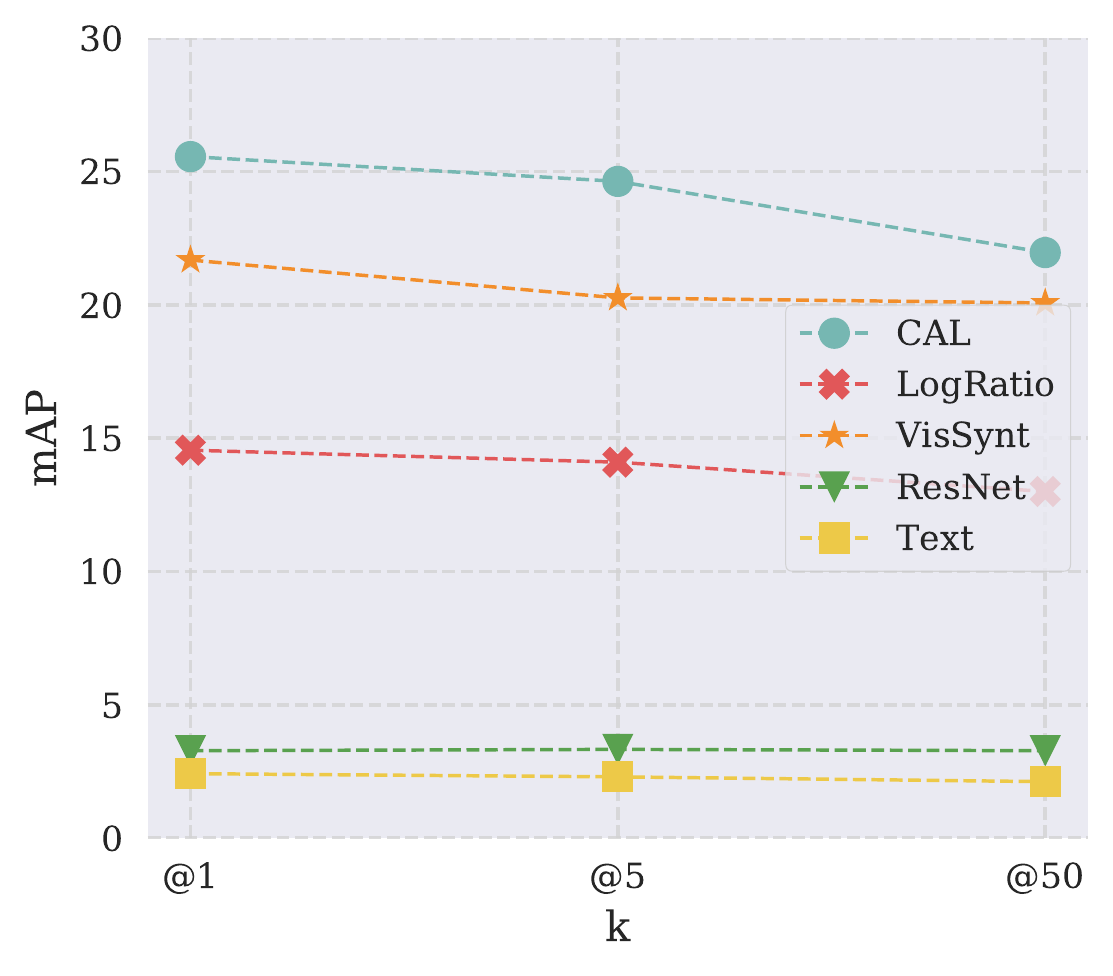}
\endminipage\hfill
\minipage{0.28\textwidth}
  \includegraphics[width=\linewidth]{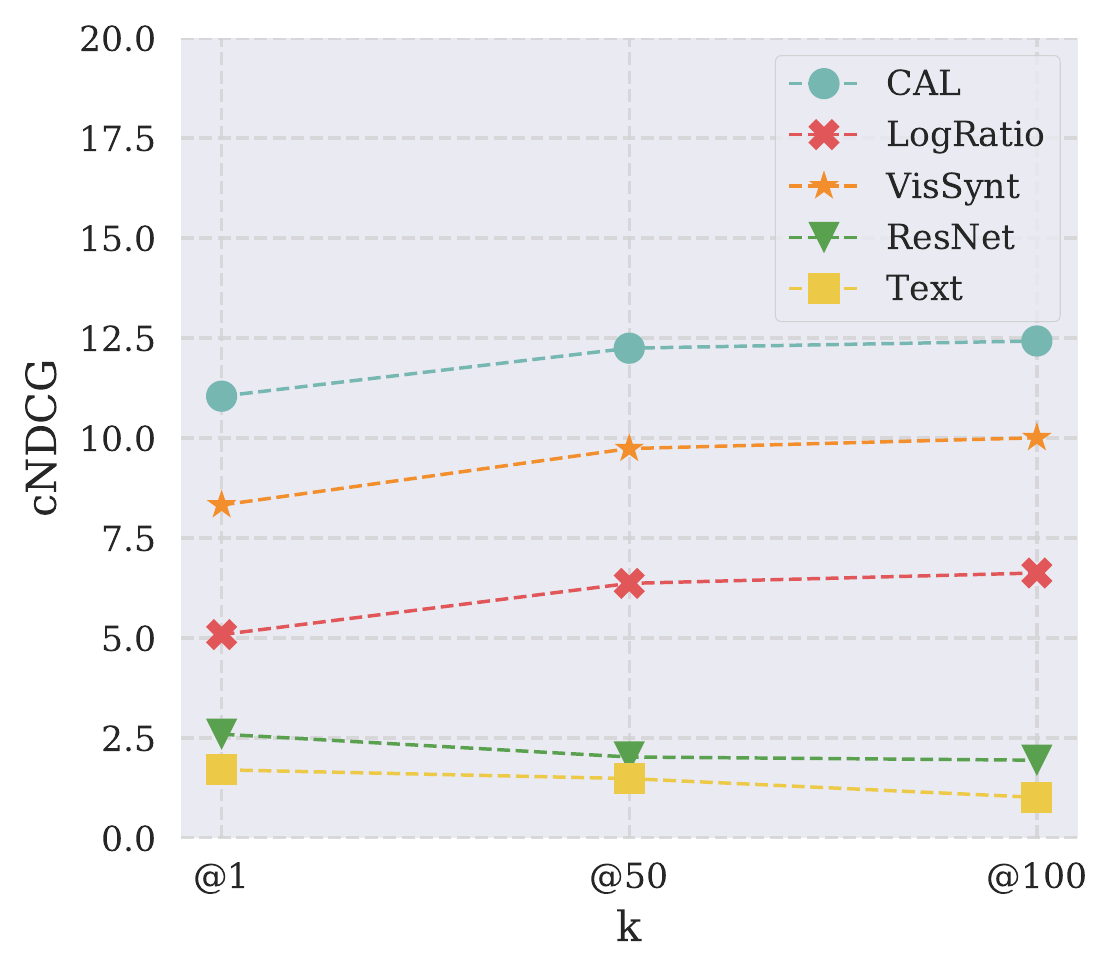}
\endminipage\hfill
\minipage{0.28\textwidth}%
  \includegraphics[width=\linewidth]{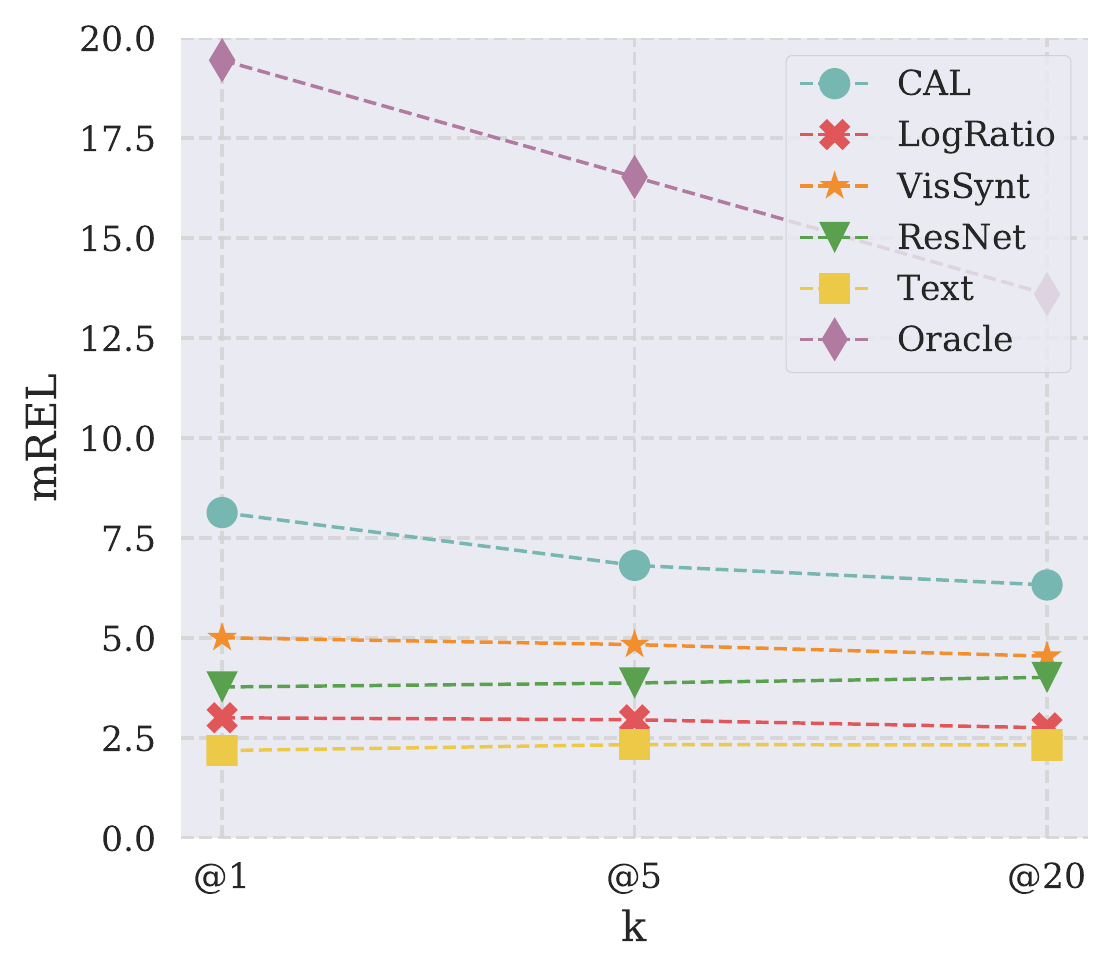}
\endminipage
\caption{Benchmarking on HICO-DET~\cite{hicodet}. Our method transfers better to HICO-DET dataset for object-interaction search.}
\label{fig:sotahico}
\end{figure*}

In the last experiment, we compare our approach to competing techniques on MS-COCO in Figure~\ref{fig:sotacoco} and HICO-DET in Figure~\ref{fig:sotahico} datasets.

As can be seen, our method outperforms the compared baselines in both datasets, and in 3 metrics. This confirms the effectiveness of composition-aware learning for object (MS-COCO) and object-interaction (HICO-DET) search. The results in HICO-DET are much lower compared to MS-COCO since 1) HICO-DET has a higher number of query images ($10k$ vs. $5k$), 2) Many queries have only a few relevant images within the gallery set (as can be seen from the oracle performance of only $0.19$ mREL in Figure~\ref{fig:sotahico}), 3) No training is conducted on HICO-DET, revealing the transfer-learning abilities of the evaluated techniques.

\begin{figure}[h]
\begin{center}
  \includegraphics[width=1.0\linewidth]{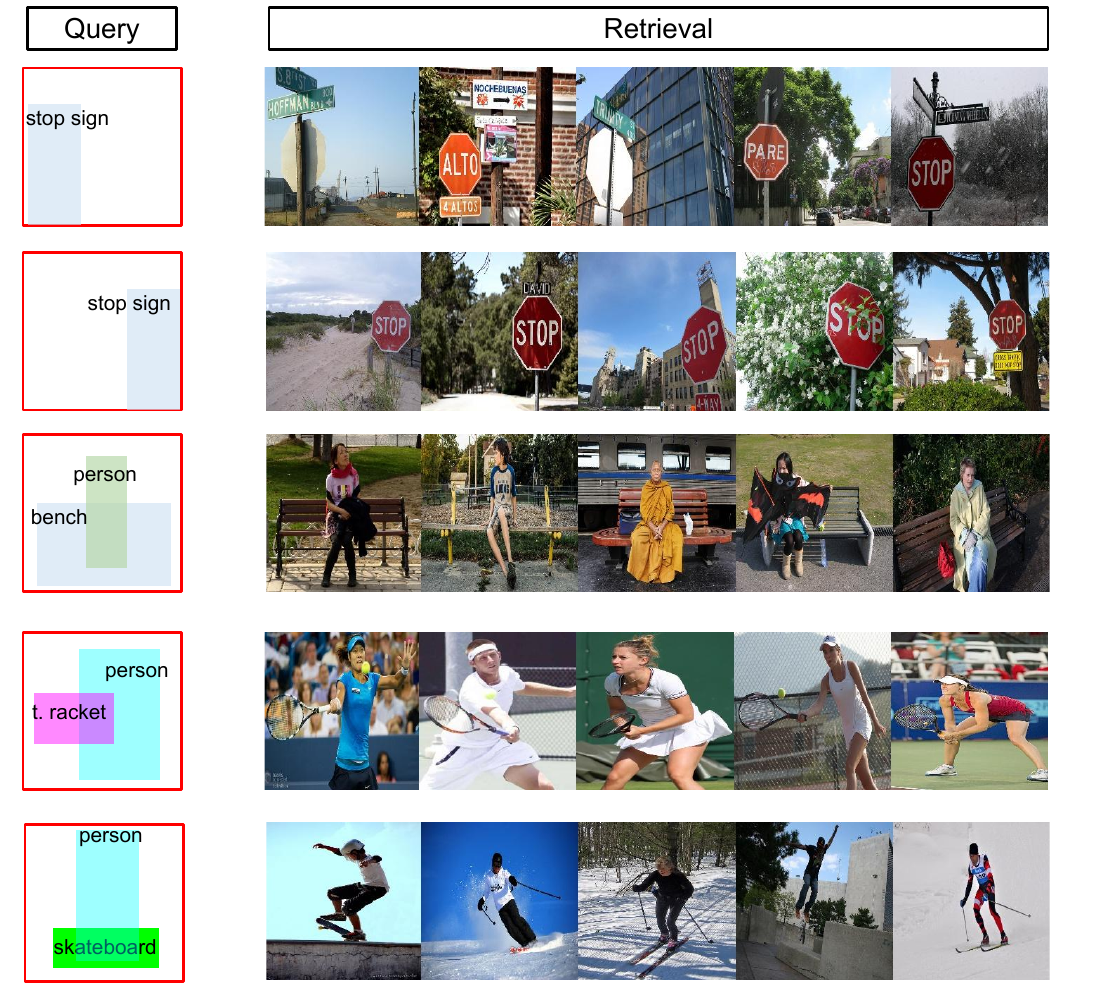}
\caption{Qualitative examples. First two rows show a single-object query, and last three rows show multi-object queries. As can be seen, our approach considers the object category, location and interaction into account while retrieving examples.}
\label{fig:qualitative}
\end{center}
\end{figure}



\partitle{Qualitative analysis.} Lastly, we showcase a few qualitative examples in Figure~\ref{fig:qualitative}. First, as a sanity check, we illustrate single object queries (stop signs). As can be seen, our method successfully retrieves images relevant to the query category and the position. Then, we illustrate some object-interaction examples, such as human-on-bench, or human-with-tennis racket, or human-on-skateboard. Our model can still generalize to such examples, meaning that compositional learning benefits the case of the object interaction. We illustrate a failure case in the last row, where our model retrieves a mix of snowboard-skateboard objects given the query of a skateboard. This indicates that our model performance can be improved by incorporating scene context, which we leave as future work.

\section{Conclusion}

In this work we tackled a structured visual search problem called compositional visual search. Our approach is based on the observation that the visual compositions are continuous-valued transformations of each other, carrying rich information. Such transformations mainly consists of the positional and categorical changes within the queries. To leverage this information, we proposed composition-aware learning which consists of the representation of the input-output transformations as well as a new loss function to learn these transformations. Our experiments reveal that defining the transformations within the visual domain is more useful than the lingual counterpart. Also, a regularized loss function is necessary to learn such transformations. Leveraging transformations with this loss function leads to an increase in the feature and data efficiency, and outperforms existing techniques on MS-COCO and HICO-DET. We hope that our work will inspire further research to incporporate structure for the structured visual search problems. 

{\small
\bibliographystyle{ieee_fullname}
\bibliography{egbib}
}

\end{document}